\pdfoutput=1

\documentclass[11pt]{article}

\usepackage{booktabs}
\usepackage[final]{acl}
\usepackage{multirow}
\usepackage{makecell}
\usepackage{appendix} 
\usepackage{microtype}
\usepackage{graphicx}
\usepackage{hyperref}
\usepackage{subfigure}
\usepackage{booktabs} 
\usepackage{graphicx}
\usepackage{subcaption}
\usepackage{caption}
\usepackage{float}
\usepackage{times}
\usepackage{latexsym}
\usepackage{marvosym}
\usepackage{amssymb}
\usepackage{amsmath}
 
\usepackage[T1]{fontenc}

\usepackage[utf8]{inputenc}

\usepackage{microtype}

\usepackage{inconsolata}

\usepackage{graphicx}

\title{HD-PiSSA: High-Rank Distributed Orthogonal Adaptation}

\author{
Yiding Wang\textsuperscript{$*$1,2}, 
Fanxu Meng\textsuperscript{$*$1,3}, 
Xuefeng Zhang\textsuperscript{1}, 
Fan Jiang\textsuperscript{1,2},  \\
\textbf{Pingzhi Tang}\textsuperscript{1,2},
\textbf{Muhan Zhang}\textsuperscript{1,3}\\
\textsuperscript{1}Institute for Artificial Intelligence, Peking University \quad 
\textsuperscript{2}Yuanpei College, Peking University \\
\textsuperscript{3}State Key Laboratory of General Artificial Intelligence, BIGAI \\
\textsuperscript{*}Equal contribution \quad 
\Letter\ Correspondence to \href{mailto:muhan@pku.edu.cn}{muhan@pku.edu.cn}
}

\begin{document}
\maketitle
\begin{abstract}

Existing parameter-efficient fine-tuning (PEFT) methods for large language models (LLMs), such as LoRA and PiSSA, constrain model updates to low-rank subspaces, limiting their expressiveness and leading to suboptimal performance on complex tasks.
To address this, we introduce \textbf{H}igh-rank \textbf{D}istributed \textbf{PiSSA} (\textbf{HD-PiSSA}\footnote{The code is available at \href{https://github.com/MuLabPKU/HD-PiSSA}{MuLabPKU/HD-PiSSA}}), a distributed PEFT approach that initializes \textbf{orthogonal adapters} across different devices and aggregates their delta updates collectively on $W$ for fine-tuning. Unlike Data Parallel LoRA or PiSSA, which maintain identical adapters across all devices, HD-PiSSA assigns different principal components of the pre-trained weights to each GPU, significantly expanding the range of update directions. This results in over 16$\times$ higher effective updated ranks than data-parallel LoRA or PiSSA when fine-tuning on 8 GPUs with the same per-device adapter rank. Empirically, we evaluate HD-PiSSA across various challenging downstream tasks, including mathematics, code generation, and multi-task learning. In the multi-task setting, HD-PiSSA achieves average gains of 10.0 absolute points (14.63\%) over LoRA and 4.98 points (6.60\%) over PiSSA across 12 benchmarks, demonstrating its benefits from the extra optimization flexibility.
\end{abstract}

\section{Introduction}

Large Language Models (LLMs) trained on general corpora have shown strong language modeling capabilities and broad generalization across various downstream applications~\cite{taori2023stanford,luo2023wizardmath}. However, adapting these general models to specific application scenarios often requires fine-tuning or additional training on domain-specific data. While fine-tuning has proven effective, Full Fine-Tuning (FFT) LLMs are usually prohibitively expensive in terms of both computation and memory~\cite{dettmers2024qlora}. The cost of updating all model parameters becomes particularly challenging when working with models containing billions of parameters. To address this challenge, Parameter-Efficient Fine-Tuning (PEFT) \cite{houlsby2019parameter} methods have been developed as more practical alternatives. These methods significantly reduce the number of trainable parameters while maintaining competitive performance, making them both efficient and scalable.

\begin{figure}[t]
\center
\includegraphics[width=0.49\textwidth]{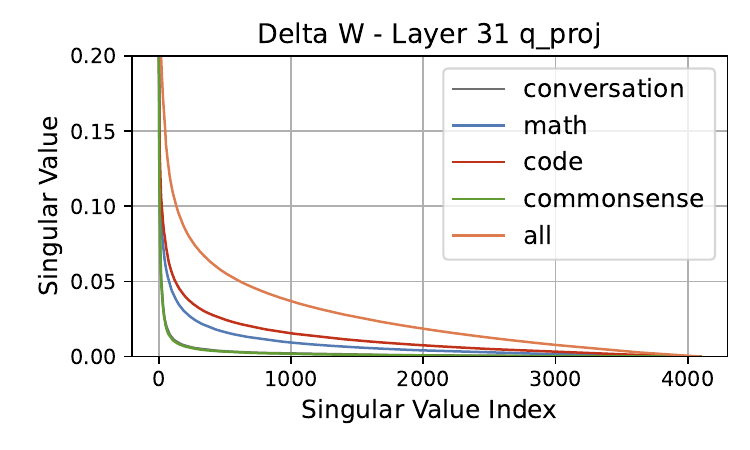}
\caption{The rank analysis of $\Delta W$ after Full Fine-Tuning (FFT) on different datasets. The magnitudes of the singular values in $S$ provide a measure of the effective updated rank. Details of this analysis are provided in Appendix~\ref{App.figure 2}.}
\label{FFT Ranks}
\end{figure}

\begin{figure*}[t]
\includegraphics[width=0.98\textwidth]{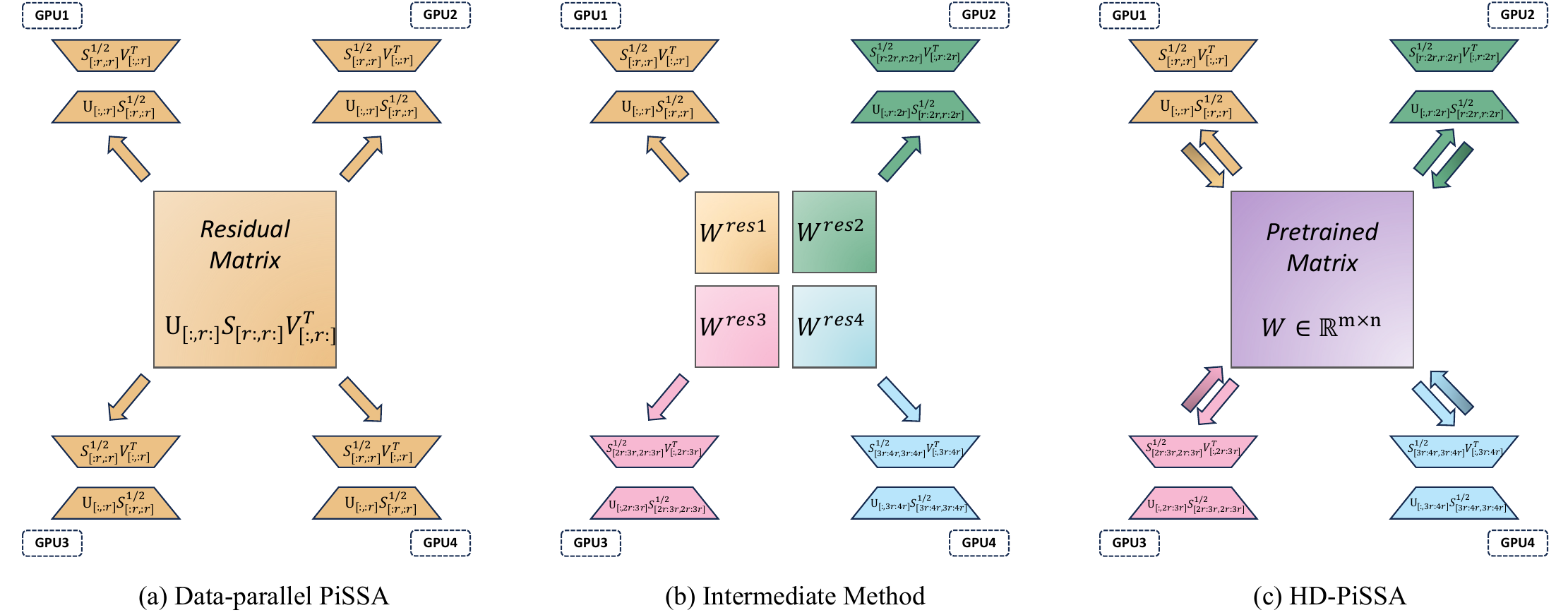}
\vspace{-5pt}
\caption{An illustration of the transition from Data-Parallel PiSSA to HD-PiSSA. 
PiSSA initializes and updates identical adapters across GPUs using only the top-\( r \) components, leading to a low-rank update limitation. A natural extension is the Intermediate Method, which increases the updated rank by distributing the top-\( Kr \) components across GPUs to construct orthogonal adapters. However, this approach introduces two key challenges: (1) How can single-step updates be performed on heterogeneous adapters? (2) Is it necessary to maintain separate copies of the residual matrix on different GPUs? HD-PiSSA addresses these challenges by introducing Direct Weight Update and a Muting Mechanism, enabling efficient, high-rank updates without the need for separate residual matrices.}
\label{Overall architecture}
\end{figure*}


Among PEFT techniques, Low-Rank Adaptation (LoRA)~\cite{hu2021lora} is widely adopted for its simplicity and efficiency. It approximates weight updates \(\Delta W\) with a low-rank decomposition \( \Delta W = AB \), based on the assumption that adaptation occurs in a low-rank subspace. This design reduces training cost and parameter count without increasing inference latency. PiSSA~\cite{meng2024pissa} improves LoRA by initializing adapters with the top singular vectors of pre-trained weights, encouraging updates to follow more informative directions. Despite these improvements, both methods share a fundamental limitation: they operate under a fixed low-rank constraint that restricts update capacity. This limitation becomes evident when adapting to complex tasks. As shown in Fig.~\ref{FFT Ranks}, Full Fine-Tuning (FFT) yields much higher update ranks on complex tasks such as code generation and math than on simpler ones like commonsense or conversation. These observations suggest that \textbf{effective adaptation to complex tasks demands higher update ranks}, highlighting the need for more expressive PEFT approaches.

To address this, we propose \textbf{HD-PiSSA}, which leverages data parallelism to break through the low-rank bottleneck. Unlike prior methods that replicate identical adapters across devices, HD-PiSSA assigns orthogonal adapters initialized from disjoint principal components to each GPU, thereby expanding the effective update subspace. Instead of updating the adapters directly, we compute and aggregate their induced updates and \textbf{apply them directly to the shared pretrained weights \(W\)}. Furthermore, to avoid maintaining separate residual matrices across devices, we introduce a \textbf{Muting Mechanism}, which enables all devices to share a single full-rank model while still benefiting from heterogeneous adapter updates. Fig.~\ref{Overall architecture} illustrates the transition from standard PiSSA to HD-PiSSA.

\begin{figure*}[hbt]
    \centering
    \begin{minipage}{0.24\textwidth}
        \centering
        \includegraphics[width=\textwidth]{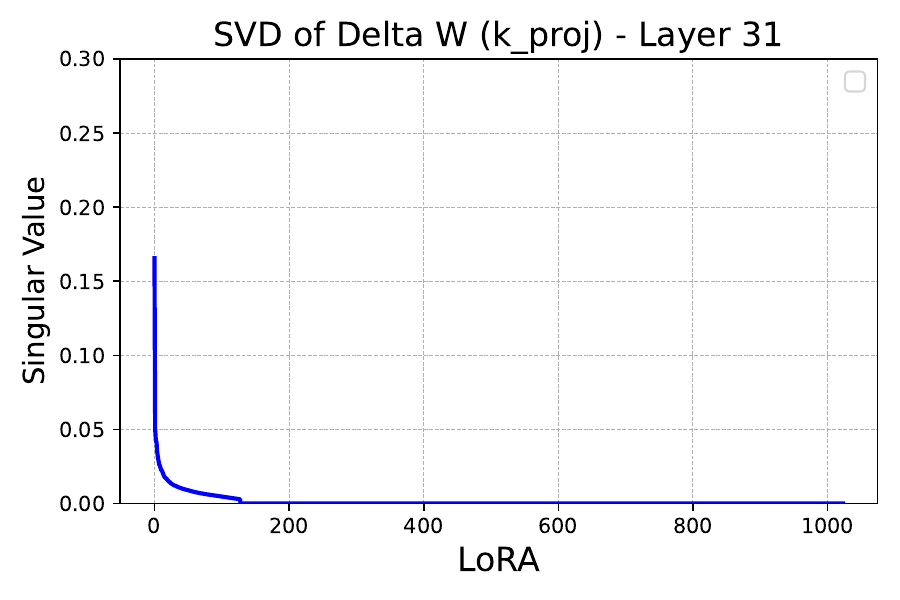}
        \vspace{-15pt}
        \captionof{subfigure}{LoRA}
    \end{minipage}
    \hfill
    \begin{minipage}{0.24\textwidth}
        \centering
        \includegraphics[width=\textwidth]{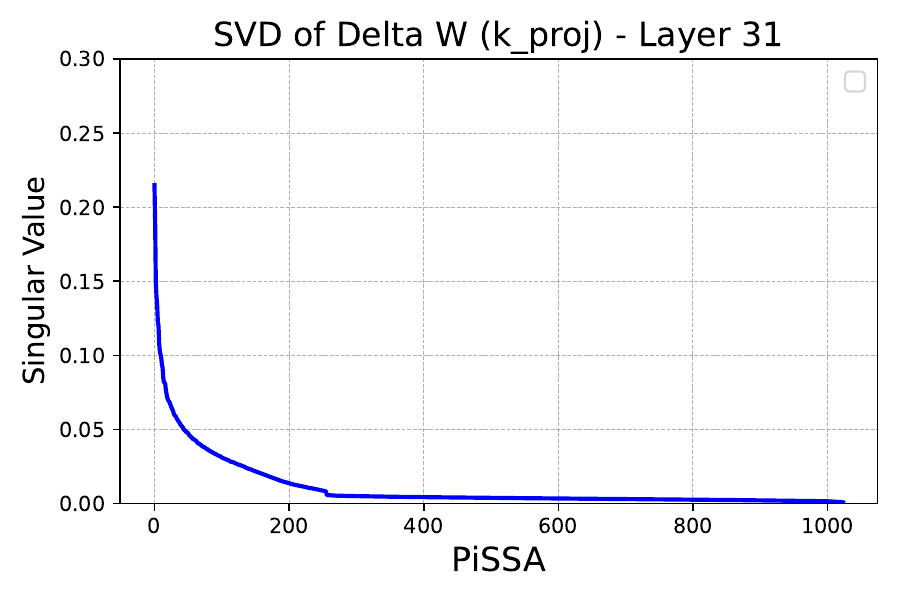}
        \vspace{-15pt}
        \captionof{subfigure}{PiSSA}
    \end{minipage}
    \hfill
    \begin{minipage}{0.24\textwidth}
        \centering
        \includegraphics[width=\textwidth]{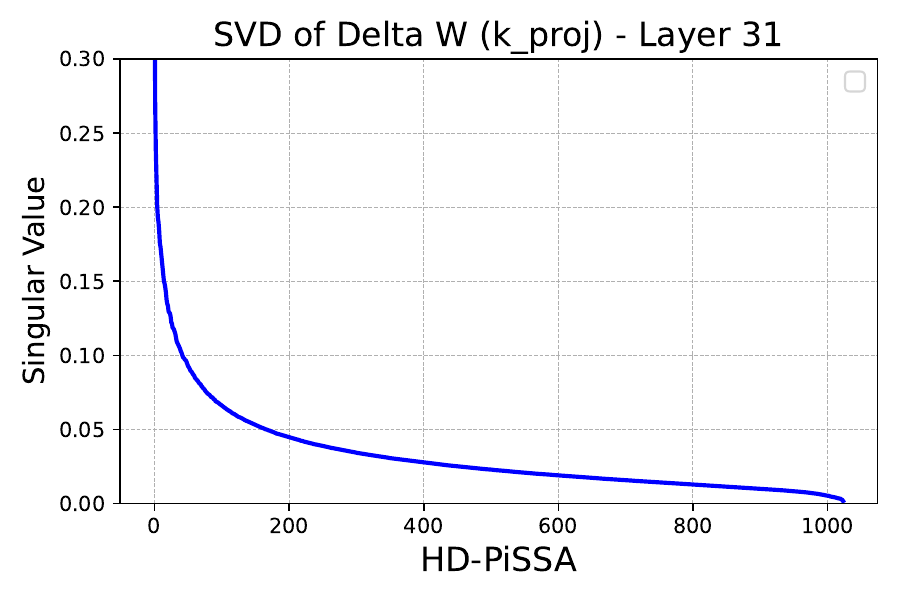}
        \vspace{-15pt}
        \captionof{subfigure}{HD-PiSSA}
    \end{minipage}
    \hfill
    \begin{minipage}{0.24\textwidth}
        \centering
        \includegraphics[width=\textwidth]{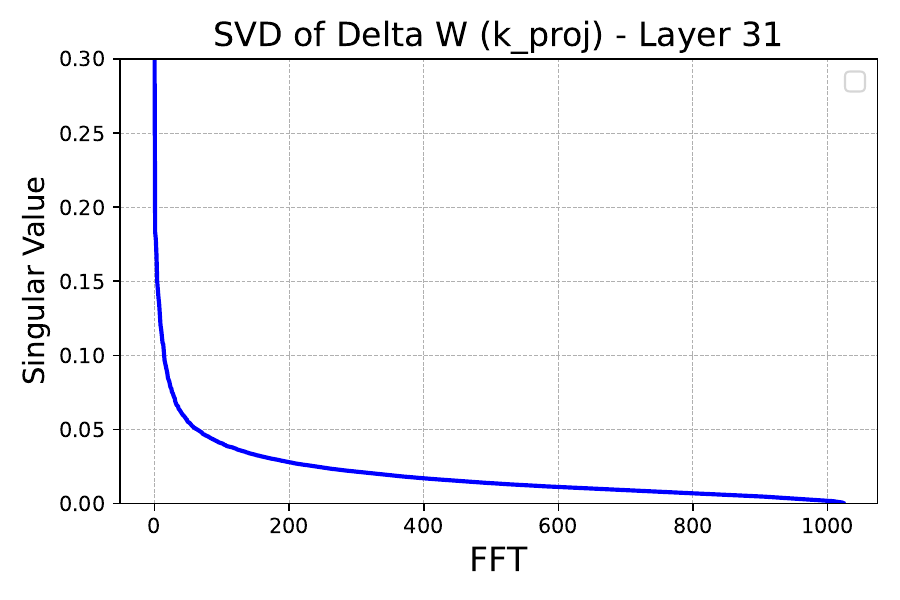}
        \vspace{-15pt}
        \captionof{subfigure}{Full Fine-Tuning}
    \end{minipage}
    \caption{SVD analysis of $\Delta W$ after fine-tuning with different methods on the same training set. The rank of $\Delta W$ is bounded by \(r\) or \(2r\) for LoRA and PiSSA. (The effective updated rank of PiSSA is also \(r\), and the appearance of \(2r\) is due to its operation of first subtracting the adapter from the pretrained weight and then merging the updates after fine-tuning.) HD-PiSSA demonstrates a similar smooth full-rank update pattern to Full Fine-Tuning. More details are included in Appendix~\ref{App.figure 3}.}
    \label{fig:comparison}
\end{figure*}

As illustrated in Fig.~\ref{fig:comparison}, HD-PiSSA overcomes the rank limitation of PiSSA and closely resembles FFT with smooth full-rank updates. We formalize this improvement through a rank analysis in Section~\ref{Ana:rank}. In addition, we evaluate the performance of HD-PiSSA on a wide range of challenging fine-tuning tasks, including mathematics, code generation, and multi-task learning. The results show that HD-PiSSA \textbf{outperforms LoRA and PiSSA while maintaining a similar number of optimizer parameters}. 
We further compare HD-PiSSA with other recent PEFT methods, like DoRA~\cite{liu2024dora} and high-rank updating methods MoRA~\cite{jiang2024mora} and HiRA~\cite{anonymous2025hira} on commonsense reasoning tasks. HD-PiSSA outperforms these baselines with a significant margin (+2.0). HD-PiSSA exhibited key characteristics of FFT, including the ability to perform high-rank or theoretically full-rank updates, underscoring its potential as an efficient and scalable alternative to Full Fine-Tuning for complex tasks.



A summary of our contributions is as follows:
\begin{itemize}
    \item We propose HD-PiSSA, a novel distributed PEFT method that enables high-rank updates under data parallel fine-tuning, achieving a learning capacity resembling FFT without additional inference latency over LoRA.
    \item HD-PiSSA consistently outperforms existing PEFT methods, including LoRA, PiSSA and other high-rank updating methods, across single-task and multi-task fine-tuning settings.
    \item We provide empirical analysis of update ranks of FFT, LoRA, PiSSA and HD-PiSSA, revealing key insights into the learning patterns and limitations of different PEFT methods.
\end{itemize}

\section{Related Works}

\subsection{Parameter-Efficient Fine-Tuning (PEFT)}

PEFT~\cite{han2024parameter} methods are designed to reduce the computational or memory cost of fine-tuning large-scale models. Broadly, PEFT methods can be categorized into four groups. The first category is \textit{partial fine-tuning}~\cite{zaken2021bitfit,xu2021raise}, which involves updating only a subset of the original model’s parameters based on a predefined selection strategy. The second category is \textit{soft prompt fine-tuning}, which introduces additional soft prompt tokens or vectors into the initial input and fine-tunes only these vectors~\cite{lester2021power,li2021prefix}. However, these methods are sensitive to input variations and initialization, which limits their effectiveness in downstream tasks. The third category is \textit{adapter-based} methods. These methods~\cite{houlsby2019parameter,pfeiffer2020adapterfusion,he2021towards} incorporate additional trainable modules into original frozen models. While these methods enhance performance, they come at the cost of increased inference-time latency.

\subsection{Low-rank Adaptation Methods}

Low-rank adaptation methods, such as LoRA~\cite{hu2021lora} and its variants, fall into the fourth category. These methods use low-rank matrices to approximate weight changes during fine-tuning and can be merged with pre-trained weights before inference. They have gained popularity due to their ability to retain the model’s original architecture while enabling efficient fine-tuning. Building on LoRA, 
O-LoRA~\cite{wang2023orthogonal} introduces orthogonal regularization to mitigate catastrophic forgetting, DeltaLoRA~\cite{zi2023delta} injects adapter-informed updates into the original weights, and DoRA~\cite{liu2024dora} decouples weight magnitude and direction.
PiSSA~\cite{meng2024pissa} initializes low-rank adapters using the principal components of the original weights and fine-tunes these components. Despite their efficiency, these methods are inherently constrained by fixed low-rank updates, which can limit their expressiveness on complex tasks.
To address this, recent work has explored high-rank adaptation. \citet{hao2024flora} interpret LoRA as a gradient compressor and propose Flora to enable high-rank updates with reduced memory. MoRA~\cite{jiang2024mora} replaces the low-rank structure with square matrices to enable full-rank updates, while HiRA~\cite{anonymous2025hira} introduces a Hadamard product mechanism to enrich the update space.
Unlike previous high-rank PEFT methods that increase capacity by modifying adapter structures on each device, HD-PiSSA leverages data-parallel training to achieve high-rank adaptation via distributed and orthogonal updates. This design retains the efficiency of low-rank methods while substantially expanding the update space.


\section{Background and Motivation}

\subsection{LoRA Basics}
In this paper, we denote the original model's weight matrix as \(W \in \mathbb{R}^{m \times n}\), where \(m\) and \(n\) represent the input and output dimensions, respectively. LoRA approximates the weight update \(\Delta W\) using a low-rank decomposition: \(\Delta W = AB\), where \(A \in \mathbb{R}^{m \times r}\) and \(B \in \mathbb{R}^{r \times n}\) with \(r \ll \min(m, n)\). The original weights \(W\) are frozen during fine-tuning, and only the low-rank matrices \(A\) and \(B\) are trainable. Typically, \(A\) is initialized with Kaiming Uniform initialization, while \(B\) is initialized to zero, resulting in an initial \(\Delta W = 0\).

\subsection{PiSSA Basics}

PiSSA builds upon the low-rank adaptation idea but introduces a more informed initialization strategy. Specifically, it first performs Singular Value Decomposition (SVD) on the original weight matrix \(W\), obtaining: \(W = U \Sigma V^\top\).
where \(U \in \mathbb{R}^{m \times d}\), \(\Sigma \in \mathbb{R}^{d \times d}\), and \(V \in \mathbb{R}^{n \times d}\), with \(d = \min(m, n)\). PiSSA then selects the top \(r\) principal components from \(U\) and \(V\) to initialize the low-rank matrices \(B\) and \(A\), respectively. That is,
\[
A = U_r \Sigma_r^{1/2}, \quad B = \Sigma_r^{1/2} V_r^\top,
\]
The original weight is then replaced by a residual form \( W_{res} = W - AB \) to preserve the model's output consistency. This initialization captures the key directions of \(W\), offering better fine-tuning performance than random initialization.

\subsection{Distributed High-Rank Adaptation}

While data parallelism is a widely used paradigm for large language model fine-tuning, existing parameter-efficient tuning methods, including PiSSA, do not fully exploit its potential. In conventional PiSSA (as shown in Fig.~\ref{Overall architecture}a), all devices fine-tune the same adapters initialized from top singular components, leading to limited update rank and underutilization of model capacity.

To address this, we observe that data parallel training naturally allows each GPU to maintain a local copy of the model and its adapters. This opens up the opportunity to assign \emph{heterogeneous}, non-overlapping adapter modules to different GPUs. By distributing distinct subsets of principal components across devices, we effectively enlarge the overall update rank without increasing memory overhead on any single GPU. This simple yet powerful insight lays the foundation for our method, HD-PiSSA, which leverages the structure of data parallelism to scale up the expressiveness of low-rank updates through adapter heterogeneity, while keeping computational and memory costs comparable to conventional approaches.

\section{Methodology}

Building on the insight that data parallelism enables heterogeneous adapters across devices, we present \textbf{HD-PiSSA} (\textbf{H}igh-rank \textbf{D}istributed \textbf{PiSSA}), which enhances update capacity while maintaining efficiency. Specifically, HD-PiSSA introduces three key modifications to standard PiSSA. 
First, we adopt \emph{orthogonal adapter initialization}: each GPU is assigned a distinct subset of principal components, enabling it to fine-tune a unique set of adapters. This increases the effective update rank without incurring extra memory costs. 
Second, we employ a \emph{direct weight update} strategy: instead of updating \(A\) and \(B\), gradients are applied directly to the original weights \(W\), while keeping the adapters fixed. Finally, we design a \emph{muting mechanism} using a learnable scalar that suppresses adapter outputs during the forward pass, enabling gradient flow without modifying model predictions or requiring subtraction of \(AB\). Together, these designs make HD-PiSSA a simple, scalable, and efficient solution for high-rank adaptation in data-parallel training.

\subsection{Orthogonal Adapters Initialization}

Assuming we are fine-tuning on \(K\) devices, HD-PiSSA begins by applying SVD to each weight matrix \( W \in \mathbb{R}^{m \times n} \):
\[
W = U S V^T,
\]
where \( U \) and \( V \) are orthogonal matrices and \( S \) contains the singular values. Instead of selecting only the top-\(r\) components as in PiSSA, we select the top-\(Kr\) components and partition them across devices. For each device \( i \in \{0, \dots, K-1\} \), we construct the adapter pair as:
\[
A_i = U_{[:,r_i:r_{i+1}]} S^{1/2}_{[r_i:r_{i+1}]} ,~ 
B_i = S^{1/2}_{[r_i:r_{i+1}]} V^\top_{[:,r_i:r_{i+1}]},
\]
where \( r_i = i \cdot r \), and \( r \) denotes the predefined rank that each device handles locally. Compared to data-parallel PiSSA, which updates only the top-rank components using the entire batch of data, HD-PiSSA initializes adapters in a way that can distribute the batch data evenly across a larger set of principal components, thereby exploring more directions and increasing the fine-tuned parameter space. Naively, different adapters imply different residual weights \( W_{res,i} = W - A_i B_i \) to maintain consistent outputs across devices. However, as we show in the Section~\ref{sec:muting}, our muting mechanism allows us to avoid this by sharing a single global weight matrix \(W\) across all devices without compromising correctness. This enables efficient distributed fine-tuning with heterogeneous adapters while preserving model consistency.

\subsection{Direct Weight Update}

With orthogonal adapters distributed across devices, their gradients lie in distinct subspaces and cannot be directly averaged, unlike in standard data-parallel LoRA or PiSSA. To address this, HD-PiSSA keeps \(A_i, B_i\) unchanged and instead translates their updates into an equivalent update on the shared weight matrix \(W\).

Specifically, we assume that each device performs forward and backward propagation and obtains the gradients \(g_{A_i}^t \in \mathbb{R}^{m \times r} \) and \(g_{B_i}^t \in \mathbb{R}^{r \times n}\) on its mini-batch data at step \(t\). Then, we are able to compute the delta update \(\Delta A_i^t \in \mathbb{R}^{m \times r}\) and \(\Delta B_i^t \in \mathbb{R}^{r \times n}\) correspondingly, where we temporarily ignore the discussion of optimizers. If we simply update \( A \) and \( B \) at each step, the process could be formalized as follows:
\[
A_i^{t+1} = A_i^t + \Delta A_i^t,  \quad  B_i^{t+1} = B_i^t + \Delta B_i^t.
\]
This leads to the following expression for the delta update \(U_i^t \in \mathbb{R}^{m \times n} \) to the product of low-rank matrices \( A_i \) and \( B_i \) at step \( t+1 \):
\[
\Delta U_i^t = \Delta A_i^t B_i^t + A_i^t \Delta B_i^t + \Delta A_i^t \Delta B_i^t,
\]
These deltas are aggregated and then applied to \(W\):
\[
W^{t+1} = W^t + \frac{1}{K} \sum_{i=0}^{K-1} \Delta U_i^t.
\]
where \(K\) is the total number of devices. 

We refer to this strategy as \textbf{Direct Weight Update}. Through Direct Weight Update, we achieve joint fine-tuning of the top-\(Kr\) principal components across different orthogonal directions of the original weight matrix, significantly increasing the effective updated rank. Additionally, since the 
\(\Delta A\) and \(\Delta B\) terms allow arbitrary rotations and translations according to data, HD-PiSSA not only updates the top-\(Kr\) components, but also enables a theoretically full-rank update. (More details are included in Section~\ref{Ana:rank}.)

\subsection{Muting Mechanism} 
\label{sec:muting}
Since we initialize orthogonal adapters on different devices, and the dot product between \(A\) and 
\(B\) in these adapters is nonzero, following PiSSA's forward propagation paradigm
\begin{align}
\label{eq:normalupdate}
Y = X(W_{res,i}+A_iB_i)
\end{align}
 would require subtracting the dot product \(AB\) from the original matrix on each device and maintaining a separate \(W_{res}\)
 for each GPU. This introduces significant operational and computational overhead, and makes it difficult to scale HD-PiSSA to larger models, as each device would need to store an independent copy of the full model. Therefore, we introduce the \textbf{muting mechanism}, which exploits the automatic differentiation mechanism and makes it feasible to store only one shared model $W$ across devices. 

Specifically, we introduce a small constant (1e-16 in our main experiments), \textbf{the Mute Scalar} \(\gamma\). During forward propagation, we multiply this scalar to eliminate the impact of different adapters on the output: 
\[
Y = X(W + \gamma A_i B_i).
\]
This operation ensures that 1) the differences among $A_i B_i$ are muted in the forward propagation so that all devices behave the same, and 2) we can maintain a single model $W$ shared across devices instead of maintaining $W_{res,i}$ for each device, thus apply the averaged update only to $W$.

Although $A_i B_i$ are muted during forward propagation, their gradients still provide directional information for update to the combined model $W$. Therefore, during backward propagation, we rescale the gradients on 
\(A\) and \(B\) to get:
\[
g_{A_i} \leftarrow \frac{1}{\gamma} \frac{\partial \mathcal{L}}{\partial A_i}, \quad g_{B_i} \leftarrow \frac{1}{\gamma} \frac{\partial \mathcal{L}}{\partial B_i},
\]
where \( \frac{\partial \mathcal{L}}{\partial A_i} \) and \( \frac{\partial \mathcal{L}}{\partial B_i} \) are the gradients of the loss \( \mathcal{L} \) with respect to the adapters \( A_i \) and \( B_i \), respectively. It is easy to prove that the obtained $g_{A_i}$ and $g_{B_i}$ \textbf{are equal to} the gradients of $A_i$ and $B_i$ in the normal forward propagation as in Equation~\ref{eq:normalupdate}. Therefore, we have recovered the actual gradients while also saving the need of different $W_{res,i}$ at the same time through this smart Muting Mechanism.

\section{Experiments}

\begin{table*}[!ht]
\caption{Accuracy comparison of LLaMA-2-7B,  LLaMA-3.1-8B, and Mistral-7B-v0.1 with LoRA, PiSSA, and HD-PiSSA on mathematics and code generation datasets. ``Params.'' denotes the per-GPU optimizer parameters, which remain the same across LoRA, PiSSA, and HD-PiSSA.}
\label{table:comparison1}
\begin{center}
\begin{small}
\begin{sc}
\resizebox{0.95\textwidth}{!}{
\begin{tabular}{l|l|rrrrr|r}
\toprule
\textbf{Model} & \textbf{Method} & \textbf{Params.} &  \textbf{GSM8K} & \textbf{MATH} & \textbf{HumanEval} & \textbf{MBPP} & \textbf{Avg.} \\
\midrule
LLaMA 2-7B
           & LoRA    & 0.59\% & 44.58 & 6.16 & 15.20 & 33.60 & 24.89\\
           & PiSSA   & 0.59\% & 49.13 & 7.46 & 18.90 & 37.10 & 28.15\\
           & \textbf{HD-PiSSA}    & 0.59\% & \textbf{52.92} & \textbf{8.38} & \textbf{21.30} & \textbf{37.30} & \textbf{29.98}\\
\midrule
LlaMA 3.1-8B 
           & LoRA    & 0.52\% & 75.66 & 25.92 & 50.60 & 66.10 & 54.57 \\
           & PiSSA   & 0.52\% & 75.74 & 27.06 & 49.40 & 66.40 & 54.65 \\
           & \textbf{HD-PiSSA}   & 0.52\% & \textbf{76.12} & \textbf{28.52} & \textbf{52.40} & \textbf{69.30} & \textbf{56.59}\\
\midrule
Mistral-7B-v0.1   
           & LoRA    & 0.58\% & 65.81 & 18.10 & 41.50 & \textbf{58.70} & 46.03 \\
           & PiSSA   & 0.58\% & 68.46 & 18.46 & 41.50 & 58.30 & 46.68 \\
           & \textbf{HD-PiSSA}   & 0.58\% & \textbf{68.68} & \textbf{19.34} & \textbf{42.10} & 57.90 & \textbf{47.01} \\
\bottomrule
\end{tabular}}
\end{sc}
\end{small}
\end{center}
\vskip -0.1in
\end{table*}

In this section, we comprehensively evaluate HD-PiSSA. We begin with single-task fine-tuning (Section~\ref{Exp:single}) on math and code datasets using LLaMA-2-7B, LLaMA3.1-8B, and Mistral-7B-v0.1, comparing against LoRA and PiSSA. In Section~\ref{Exp:multi}, we extend to a more challenging multi-task setting that includes math, code, conversation, and commonsense reasoning. We also benchmark HD-PiSSA under the popular commonsense evaluation setup from~\citet{liu2024dora}, comparing it to high-rank PEFT baselines like MoRA, HiRA, and DoRA. Lastly, we examine HD-PiSSA’s robustness across different numbers of devices (Section~\ref{Exp:devices}) and ranks (Section~\ref{Exp:ranks}).

\subsection{Performance for Single-Task Fine-Tuning}
\label{Exp:single}

As shown in Fig.~\ref{FFT Ranks}, we use math and code datasets to represent complex single-task datasets. Specifically, we use MetaMathQA~\cite{yu2023metamath} to fine-tune mathematical problem-solving capability and then use GSM8K~\cite{cobbe2021training} and MATH datasets for evaluation. We use CodeFeedback~\cite{zheng2024opencodeinterpreter} to fine-tune and evaluate coding proficiency using the HumanEval~\cite{chen2021evaluating} and MBPP~\cite{austin2021program} datasets. We evaluate the fine-tuning performance of HD-PiSSA, compared with LoRA and PiSSA on three models: LLaMA-2-7B, LLaMA-3.1-8B, and Mistral-7B-v0.1. These three models cover common model architectures, with LLaMA-3.1-8B and Mistral-7B-v0.1 utilizing Group Query Attention (GQA)~\cite{ainslie2023gqa}. To provide a fair comparison, we uniformly set the rank \(r\) to 16 and keep the hyper-parameters consistent. More details are included in Appendix~\ref{App.Exp1}.
For each model, we fine-tune with each method on two training datasets separately for one epoch and then evaluate their performance on the corresponding test sets. Results are reported in Table~\ref{table:comparison1}.

HD-PiSSA consistently outperforms LoRA and PiSSA on complex tasks that demand higher adaptability and rank-update, such as mathematics and code generation. As shown in Table~\ref{table:comparison1}, HD-PiSSA achieves the best results across all models on the GSM8K, MATH, and HumanEval datasets. On the GSM8K test set for LLaMA-2-7B, HD-PiSSA surpasses PiSSA and LoRA by 3.79\% and 8.34\%, respectively. Similarly, on HumanEval and MBPP for LLaMA-3.1-8B, HD-PiSSA exceeds PiSSA by nearly 3\% in both cases. These improvements demonstrate that, compared to PiSSA and LoRA, HD-PiSSA effectively utilizes a similar number of optimizer parameters per device while expanding the set of update directions. Its higher updated rank and flexible parameter space significantly enhance learning capacity, particularly for challenging fine-tuning scenarios.

\subsection{Performance for Multi-Task Fine-Tuning}\label{Exp:multi}

\begin{figure}[h]
\includegraphics[width=0.44\textwidth]{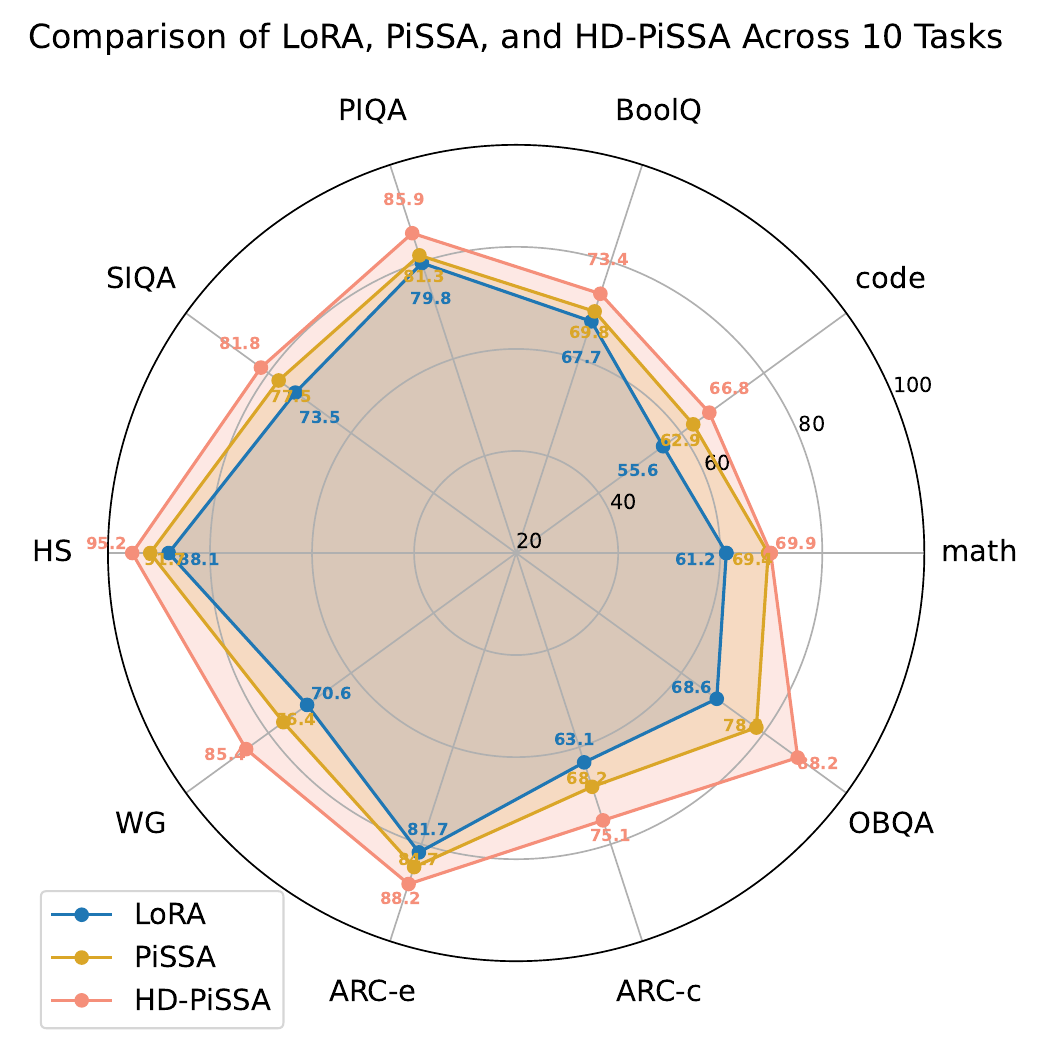}
\caption{HD-PiSSA outperforms LoRA and PiSSA in multi-task fine-tuning setting. The absolute values of two math evaluation datasets (GSM8K, MATH) and two code datasets (MBPP, HumanEval) are summed up to keep value range consistency. }
\label{Fig:Multi-task}
\end{figure}

\begin{table*}[ht]
\caption{Accuracy comparison of LLaMA-2 7B with various PEFT methods on eight commonsense reasoning datasets. Results for ChatGPT, LoRA, and DoRA are sourced from \citet{liu2024dora}, and results for MoRA and HiRA are sourced from \citet{anonymous2025hira}.}
\vskip 0.1in
\setlength{\tabcolsep}{1.2mm}
\centering
\begin{small}
\begin{sc}
\resizebox{0.95\textwidth}{!}{
\begin{tabular}{c|c|ccccccccc|cc}
\toprule
\textbf{Model} & \textbf{Method} & \textbf{Params} & \textbf{BoolQ} & \textbf{PIQA}&\textbf{SIQA}& \makecell{\textbf{Hella}\\\textbf{Swag}} & \makecell{\textbf{Wino}\\\textbf{Grande}}& \textbf{ARC-e} & \textbf{ARC-c} & \textbf{OBQA} & \textbf{Avg.} \\ \hline
ChatGPT & - & - & 73.1 & 85.4 & 68.5 & 78.5 & 66.1 & 89.8 & 79.9 & 74.8 & 77.0 \\ \hline

\multirow{4}{*}{LLaMA-2-7B}  
& LoRA & 0.83\% & 69.8& 79.9& 79.5& 83.6& 82.6& 79.8& 64.7& 81.0& 77.6 \\ 
& DoRA & 0.84\% & 71.8 & \underline{83.7}& 76.0& \underline{89.1}& 82.6& 83.7& 68.2& 82.4& 79.7 \\  
& MoRA & 0.82\% & \underline{72.2} & 80.8 & \underline{79.5} & 29.0 & 80.2 & 85.3 & 71.4 & 81.2  & 72.5 \\
& HiRA & 0.83\% & 71.2 & 83.4 & \underline{79.5} & 88.1 & \underline{84.0} & \underline{86.7} & \textbf{73.8} & \textbf{84.6} & \underline{81.4} \\
& \textbf{HD-PiSSA} & 0.83\% & \textbf{73.7} & \textbf{84.6} & \textbf{81.1} & \textbf{94.0} & \textbf{84.7} & \textbf{87.5} & \underline{72.4} & \underline{84.2} & \textbf{83.4} \\
\bottomrule
\end{tabular}}
\end{sc}
\end{small}
\label{tab:llama_commonsense}
\vskip -0.1in
\end{table*}

Based on our earlier analysis, we hypothesized that HD-PiSSA would demonstrate even stronger performance in more challenging multi-task fine-tuning settings due to its ability to overcome the rank limitations of LoRA and PiSSA and update parameters in a higher-dimensional space. 
To evaluate this hypothesis, we construct a multi-task fine-tuning dataset by combining data from four domains: mathematics, code generation, commonsense reasoning, and conversation. Specifically, we use MetaMathQA, CodeFeedback, WizardLM-Evol-Instruct~\cite{xu2024wizardlm}, and eight widely used commonsense reasoning datasets~\cite{liu2024dora} to build a cross-domain dataset (see Appendix~\ref{App.dataset}). This combined dataset represents a substantially more complex fine-tuning scenario, with a diverse range of tasks requiring broad adaptability and higher learning capacity. For all PEFT methods, we use adapters with 128 ranks to fit the increased difficulty of the multi-task setting. The models are fine-tuned on LLaMA-2-7B for one epoch, and evaluations are conducted using the same benchmarks as the single-task experiments in Section~\ref{Exp:single}, as well as the test sets of the eight commonsense reasoning datasets. 

As shown in Fig.~\ref{Fig:Multi-task}, HD-PiSSA consistently demonstrates significant improvements over both PiSSA and LoRA across mathematics, programming, and commonsense reasoning tasks. Especially on some commonsense datasets, such as OpenBookQA (78.2\%/88.2\%) and WinoGrande (76.4\%/85.4\%), it achieves nearly a 10\% performance improvement. This indicates that HD-PiSSA's enhanced updated rank flexibility allows it to capture and retain a greater volume of task-specific knowledge, which is essential for adapting to the diverse and complex data distributions present in multi-task fine-tuning.

We further evaluate HD-PiSSA on a widely adopted multi-task commonsense reasoning setting~\cite{liu2024dora}, comparing it against high-rank PEFT methods such as MoRA, HiRA, and DoRA. Training is conducted on eight standard datasets (BoolQ, PIQA, SIQA, HellaSwag, WinoGrande, ARC-c, ARC-e, and OBQA), and evaluated on their respective test sets. To ensure fairness, we match the number of trainable parameters by tuning only the \(q\_proj, k\_proj, v\_proj\) and \(up, down\) matrices for three epochs. We report results from the final model without validation-based checkpointing. As shown in Table~\ref{tab:llama_commonsense}, HD-PiSSA achieves a 2.0\% average gain over the previous state-of-the-art (HiRA), and consistently outperforms other high-rank approaches.

\section{Ablation Study}

\subsection{Experiments on Different Numbers of Devices}
\label{Exp:devices}

To evaluate the distributed nature of HD-PiSSA, we conducted ablation experiments using different numbers of GPUs (2, 4, and 8) during training. All experiments were performed on the same code dataset for 3 epochs, with evaluations on HumanEval (H) and MBPP (M) conducted every 500 steps. The total batch size was fixed at 128 for all configurations to ensure comparability. Other hyperparameters remained unchanged.

The motivation behind this experiment is rooted in HD-PiSSA’s design: as the number of devices increases, so does the number of orthogonal adapters and their corresponding data partitions. This enhanced modularization is expected to improve both training efficiency and performance.

\begin{figure}[t]
\includegraphics[width=0.48\textwidth]{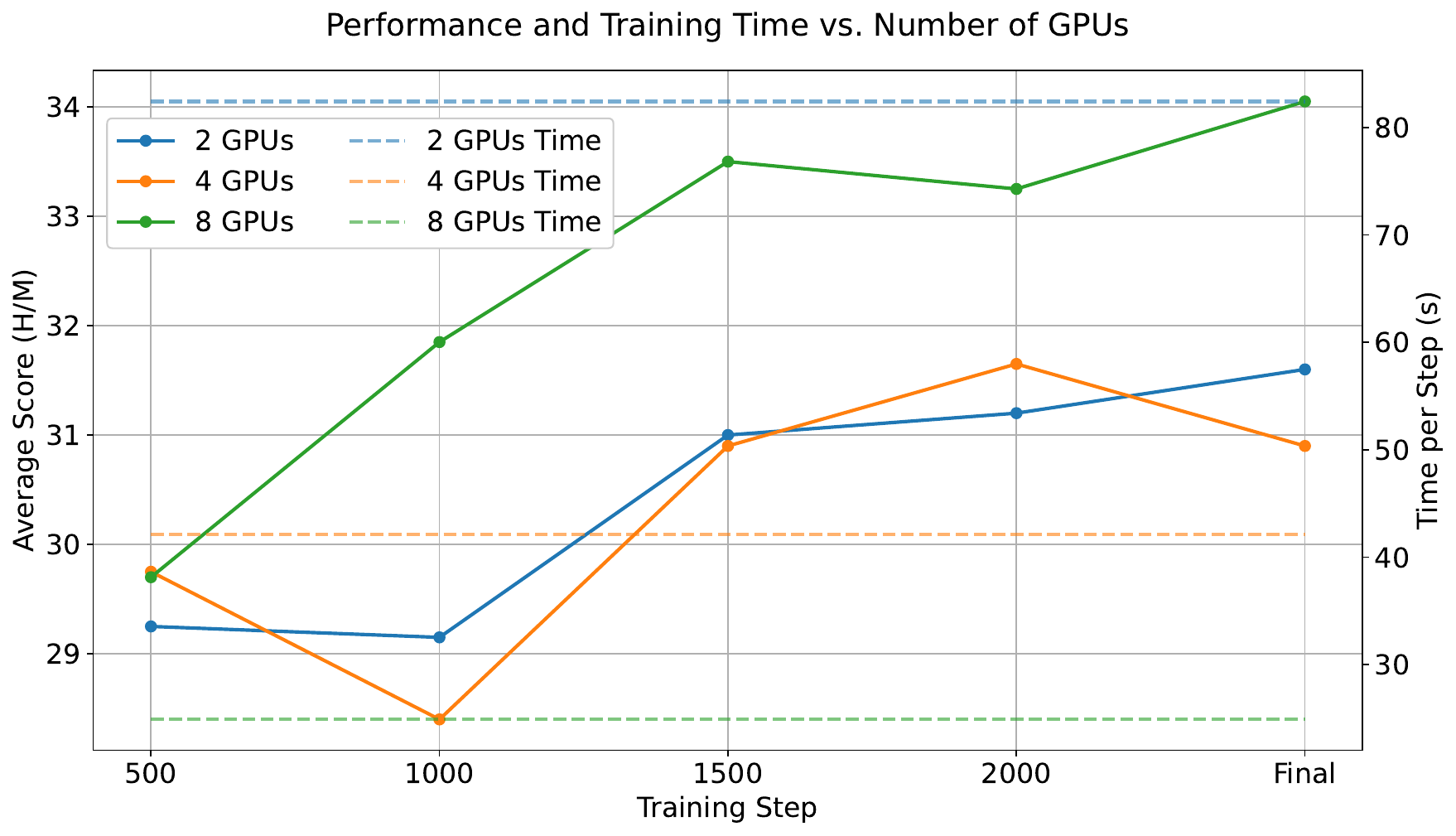}
\caption{Average performance on code generation task and training time with different numbers of devices during distributed fine-tuning. }
\label{Fig:num_devices}
\end{figure}

As shown in Fig.~\ref{Fig:num_devices}, increasing the number of GPUs significantly reduces training time per step (from 82.44s to 24.88s) while also leading to improved final performance. With 8 devices, HD-PiSSA achieves the best average score (34.05), showing that the distributed modularity inherent in the method benefits both scalability and quality of learned representations. This confirms the strength of HD-PiSSA in leveraging increased computational resources effectively.

\subsection{Experiments on Various Ranks}
\label{Exp:ranks}
\begin{figure*}[ht]
    \centering
    \begin{minipage}{0.46\textwidth}
        \centering
        \includegraphics[width=\textwidth]{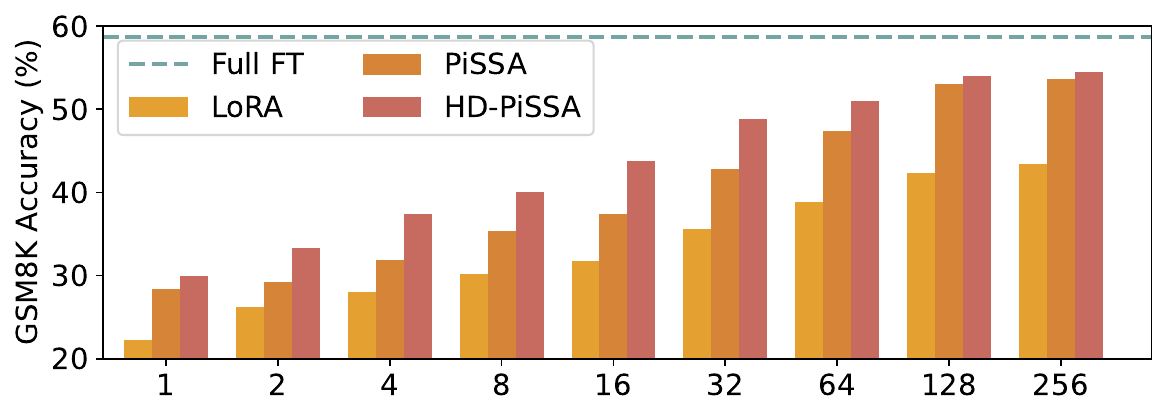}
        \captionof{subfigure}{Accuracy on GSM8K under various ranks.}
    \end{minipage}
    \hfill
    \begin{minipage}{0.46\textwidth}
        \centering
        \includegraphics[width=\textwidth]{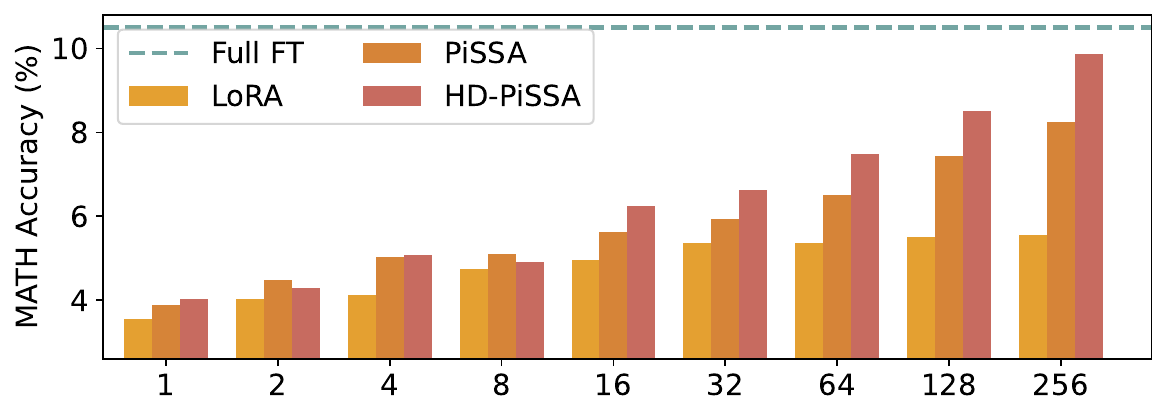}
        \captionof{subfigure}{Accuracy on MATH under various ranks.}
    \end{minipage}
    \vspace{-5pt}
    \caption{Performance comparison of LoRA, PiSSA, and HD-PiSSA across different adapter ranks (1 to 256). The dashed line represents the accuracy of Full Fine-Tuning (FFT) as a reference. The results for LoRA and PiSSA within 1-128 rank range are taken from \citet{meng2024pissa}, with FFT performed using the Float32 data type rather than the BF16 data type used in their original FFT results.}
    \label{fig:ranks}
\end{figure*}
We study how different ranks affect HD-PiSSA’s performance using LLaMA-2-7B and 100K samples from MetaMathQA. Evaluations on GSM8K and MATH (Fig.~\ref{fig:ranks}) show that HD-PiSSA generally outperforms LoRA and PiSSA, especially at higher ranks. On MATH, HD-PiSSA approaches Full Fine-Tuning performance with only a 0.62\% gap at rank 256.

When comparing HD-PiSSA with PiSSA, we observe distinct patterns across datasets. On GSM8K, HD-PiSSA performs better than PiSSA in the 2-32 rank range. On MATH, however, HD-PiSSA’s advantage continues to grow beyond rank 8. This aligns with the fact that MATH is a more complex dataset requiring more updates to learn effectively, while GSM8K is simpler and benefits from fewer updates. Consequently, the difference between the methods is less pronounced at lower ranks on MATH. These results show that HD-PiSSA can achieve near-FFT performance with much lower ranks than PiSSA, demonstrating its efficiency in handling complex tasks with fewer updates.

\subsection{Performances with Different Mute Scalars}
\label{Abl:Mute Scalar}

\begin{figure}[ht]
\centering
\includegraphics[width=0.46\textwidth]{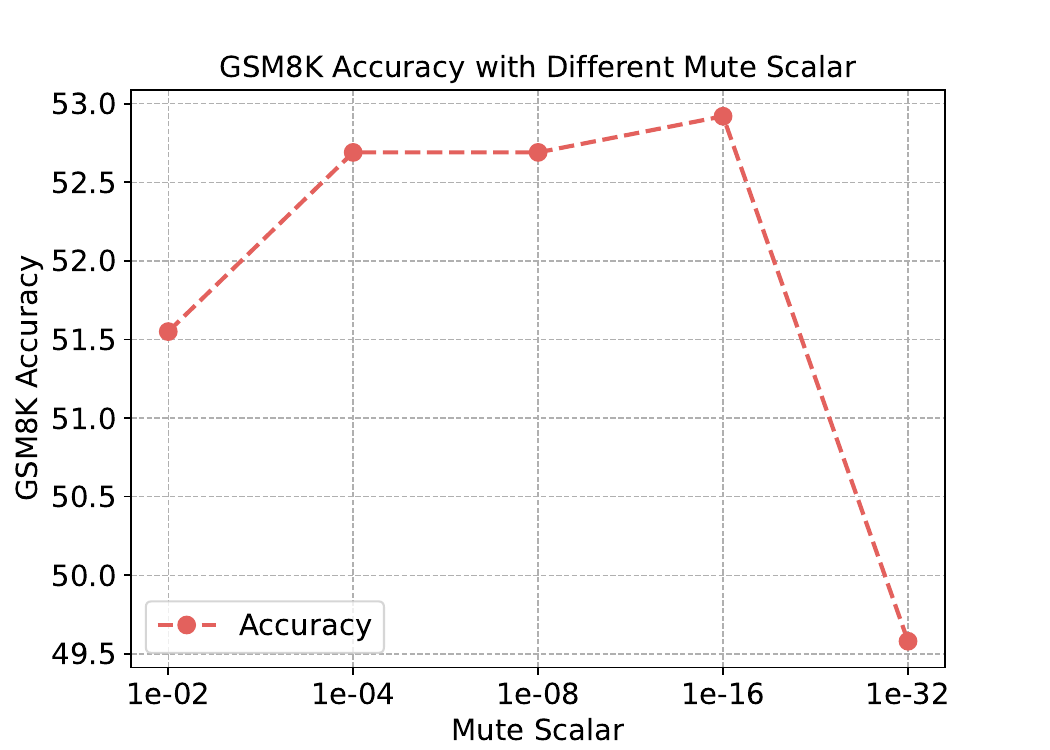}
\caption{Performances on GSM8K with different mute scalars.}
\label{fig:Abl:Mute Scalar}
\end{figure}

We also evaluate the impact of different mute scalar values on performance. We tested several mute scalar values, including \(1 \times 10^{-4}\), \(1 \times 10^{-8}\), \(1 \times 10^{-16}\), \(1 \times 10^{-2}\), and \(1 \times 10^{-32}\). Specifically, we finetune LLaMA-2-7B with these mute scalars on MetaMathQA for one epoch, then evaluate the performance on the GSM8K test set. As shown in Fig.~\ref{fig:Abl:Mute Scalar}, performance remained consistent for mute scalar values of \(1 \times 10^{-4}\), \(1 \times 10^{-8}\), and \(1 \times 10^{-16}\). However, when the mute scalar was set to \(1 \times 10^{-2}\) or \(1 \times 10^{-32}\), performance decreased significantly. 

The likely cause of this behavior is that if the mute scalar is relatively large, the adapter's contributions during forward propagation are not fully eliminated, which affects the output \(y\). On the other hand, setting the scalar too small might result in reaching the lower limit of float32 precision, causing inaccuracies in gradient updates.

\subsection{More Ablation Study}
We conducted future ablation studies to verify the effectiveness of Direct Weight Update (DWU) and Orthogonal Adapter Initialization (OAI). Specifically, we included: (1) \textbf{LoRA init. (with DWU)}: A variant using standard LoRA initialization with the Direct Weight Update technique. (2) \textbf{Top-Rank HD-PiSSA (w/o OAI)}: HD-PiSSA without orthogonal adapters, using PiSSA adapter instead. All variants were fine-tuned on the code dataset for 3 epochs. Evaluation was conducted on HumanEval and MBPP (abbreviated as H\&M in the table) every 500 steps for detailed training-time insights. The total batch size and training configuration were kept consistent across methods.

\begin{table*}[ht]
\centering
\caption{Ablation study on Direct Weight Update (DWU) and Orthogonal Adapter Initialization (OAI). Results shown as HumanEval / MBPP / Average (H/M/A).}
\resizebox{\textwidth}{!}{%
\begin{tabular}{l|c|c|c|c|c|c}
\toprule
\textbf{Components / Steps} & \textbf{Avg. Loss} & \textbf{500} & \textbf{1000} & \textbf{1500} & \textbf{2000} & \textbf{Final (2323)} \\
\midrule
LoRA init. (w. DWU) & 0.6035 & 21.3/37.3/\underline{29.3} & 23.2/34.9/29.05 & 25.6/40.2/32.9 & 27.4/39.4/\textbf{33.4} & 26.2/39.9/33.05 \\
Top-Rank HD-PiSSA (w/o OAI) & \underline{0.5984} & 20.1/37.3/28.7 & 23.8/37.0/\underline{30.4} & 25.6/41.0/\underline{33.3} & 26.8/37.6/32.2 & 28.0/38.4/\underline{33.2} \\
HD-PiSSA & \textbf{0.5854} & 19.5/39.9/\textbf{29.7} & 25.6/38.1/\textbf{31.85} & 26.8/40.2/\textbf{33.5} & 26.8/39.7/\underline{33.25} & 27.4/40.7/\textbf{34.05} \\
\bottomrule
\end{tabular}
}
\label{tab:ablation-dwu-oai}
\end{table*}

Analyzing the results in Table~\ref{tab:ablation-dwu-oai}, we conclude that HD-PiSSA's superior performance is due to the synergy of DWU and OAI. First, the use of DWU clearly provides a measurable gain, as seen in the improved performance of LoRA+DWU compared to standard baselines. This aligns with findings in prior work such as AdaLoRA. However, even with DWU, both LoRA and PiSSA-based variants lag behind HD-PiSSA in terms of average loss and final evaluation metrics.

We also test HD-PiSSA's training overhead compared with PiSSA, and found that HD-PiSSA's time and memory costs slightly increase. Details in the Appendix~\ref{appendix:speed}

\section{Analysis of Effective Update Rank}
\label{Ana:rank}

\label{Ana:rank}

We provide an intuitive explanation for why HD-PiSSA enables richer and higher-rank updates than traditional PEFT methods like LoRA and PiSSA.

\paragraph{Rank Expansion via Orthogonal Adapters.}
Each device hosts an orthogonal adapter of rank \(r\), and its update takes the form:
\[
\Delta W^t = \frac{1}{K} \sum_{i=0}^{K-1} \left( \Delta A_i^t B_i + A_i \Delta B_i^t + \Delta A_i^t \Delta B_i^t \right).
\]
Assuming the adapter matrices \(A_i\) and \(B_i\) are orthogonal and uncorrelated, the first-order terms alone contribute an approximate update rank of \(2Kr\), significantly higher than the single-device LoRA/PiSSA rank of \(r\).

\begin{figure}[t]
\centering
\includegraphics[width=0.48\textwidth]{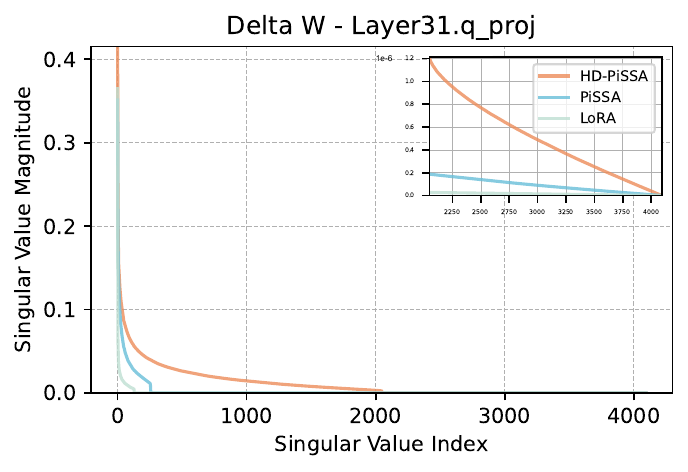}
\caption{Comparison of singular value magnitudes for the update matrices of LoRA ($r=128$), PiSSA ($r=128$), and HD-PiSSA ($r=128$). The sub-figure in the top-right corner provides a magnified view of the singular values for indices beyond $2Kr$. This highlights that the singular values for HD-PiSSA remain 1 to 2 orders of magnitude larger than those of LoRA and PiSSA in this high-rank region, empirically supporting that HD-PiSSA achieves update rank beyond $2Kr$.}
\label{fig:rank_expansion}
\end{figure}

\paragraph{Going Beyond 2\(\textbf{Kr}\).}
The second-order term \(\Delta A_i^t \Delta B_i^t\), though small, introduces updates in additional directions not captured by the main components. Since \(\Delta A\) and \(\Delta B\) evolve during training, they can rotate the updates into less-explored regions of the parameter space. As a result, HD-PiSSA’s updates extend beyond the top-\(2Kr\) components. This is empirically validated in Fig.~\ref{fig:rank_expansion}, where the singular values of HD-PiSSA remain 1–2 orders of magnitude larger than LoRA and PiSSA beyond the \(2Kr\) threshold—indicating stronger coverage of the parameter space and more expressive adaptation capability.

\section{Conclusion}

We propose \textbf{HD-PiSSA}, a parameter-efficient fine-tuning method that overcomes the rank limitations of LoRA and PiSSA while maintaining the same memory cost. By distributing orthogonal adapters across GPUs and applying Direct Weight Update, HD-PiSSA significantly expands the update rank and achieves performance close to Full Fine-Tuning. The Muting Mechanism further improves our method by eliminating the need to maintain the residual matrices as PiSSA. Experiments on mathematics, code generation, and multi-task learning show consistent improvements over existing PEFT methods. Ablation studies further validate the effectiveness of our design choices. HD-PiSSA provides a scalable and efficient fine-tuning strategy for LLMs, making it a strong alternative to Full Fine-Tuning.

\section{Limitations}

HD-PiSSA requires multi-GPU distributed setups to fully realize its benefits, which may restrict its applicability in resource-constrained environments. Besides, because HD-PiSSA aggregates all the gradient updates on the Pretrained Matrix W, its performance decreases when the base model is loaded at lower precision. Additionally, while demonstrating strong results on various tasks, its generalization to broader domains and modalities remains to be validated. We also plan to conduct more rigorous theoretical analysis on the effective update rank as part of future work.



\bibliography{custom}
\clearpage
\appendix

\section{Training Details}

\subsection{Computing Infrastructure}

All experiments were conducted on NVIDIA A100 GPUs with 40GB or 80GB memory configurations. Running the full set of main experiments, including all primary tables, required approximately 21 days using 8 GPUs in parallel.

\subsection{Implementation details of the experiment in Figure 1}
\label{App.figure 2}

\noindent \textbf{Model:} LLaMA-2-7B \\
\textbf{Optimizer:} AdamW, learning rate = 2e-5, warm-up ratio = 0.03 \\
\textbf{Epochs:} 1 \\
\textbf{Batch size:} 128 \\
\textbf{Visualized layer:} Layer 31, Query projection (\texttt{q\_proj}) 

\begin{table*}[htbp]
\caption{Details of the datasets used for training and evaluation.}
\centering
\begin{tabular}{|l|l|}
\hline
\textbf{Dataset} & \textbf{Description} \\ \hline
\textbf{math} & MetaMathQA \\ \hline
\textbf{code} & CodeFeedback \\ \hline
\textbf{commonsense} & BoolQ, PIQA, SIQA, HellaSwag, WinoGrande, Arc-easy, Arc-challenge, OBQA \\ \hline
\textbf{conversation} & WizardLM-Evol-Instruct \\ \hline
\end{tabular}

\end{table*}

\subsection{Implementation details of the experiment in Figure 3}
\label{App.figure 3}

\noindent \textbf{Model:} LLaMA-3.1-8B \\
\textbf{Training set:} MetaMathQA \\
\textbf{Optimizer:} AdamW, learning rate = 2e-5, warm-up ratio = 0.03 \\
\textbf{Epochs:} 1 \\
\textbf{Batch size:} 128 \\
\textbf{Ranks Per Device For LoRA, PiSSA and HD-PiSSA:} 128 \\
\textbf{Visualized layer:} Layer 31, Key projection (\texttt{k\_proj}) 

\subsection {Training Setting in Main Experiments}
\label{App.Exp1}

Details are presented in Table~\ref{tab:training_details}. Results are obtained from a single time running. Our evaluation codebase follows PiSSA's open-source evaluation codebase.

\begin{table*}[htbp]
\caption{Hyperparameter configurations used in all experiments.}
\centering
\begin{tabular}{|l|l|}
\hline
\textbf{Hyperparameter} & \textbf{Value} \\ \hline
\textbf{Optimizer} & AdamW \\ 
\textbf{Batch size} & 128 \\ 
\textbf{Learning rate (LLaMA models)} & \(2 \times 10^{-5}\) \\ 
\textbf{Learning rate (Mistral-7B-v0.1)} & \(5 \times 10^{-6}\) \\ 
\textbf{Scheduler} & Cosine annealing \\ 
\textbf{Warmup ratio} & 0.03 \\ 
\textbf{Weight decay} & 0.0 \\ 
\textbf{LoRA alpha} & Equal to LoRA rank \\ 
\textbf{LoRA dropout} & 0 \\ 
\textbf{Adapter insertion} & Applied to all linear layers \\ 
\textbf{Base Model Precision} & float32 \\ 
\bottomrule
\end{tabular}

\label{tab:training_details}
\end{table*}

\subsection{Hyperparameters Configuration in Commonsense Reasoning Experiment}
\label{App.hyper-cs}
\begin{table}[H]
\centering
\caption{Hyperparameters for HD-PiSSA}
\begin{tabular}{|l|l|}
\hline
\textbf{Hyperparameter} & \textbf{HD-PiSSA} \\ \hline
Rank \(r\) & 32 \\ 
\(\alpha\) & 64 \\
Dropout & 0.05 \\ 
Optimizer & AdamW \\ 
Learning Rate (LR) & 2e-5 \\ 
LR Scheduler & Linear \\ 
Batch Size & 16 \\ 
Warmup Steps & 100 \\
Epochs & 1 \\ 
Matrices & Q, K, V, Up, Down \\ \hline
\end{tabular}
\end{table}

\section{Multi-task Fine-Tuning Dataset}

The details of the multi-task dataset are provided in Table~\ref{tab:multi-dataset}.

\label{App.dataset}
\begin{table*}[t]
\caption{Details of the datasets used in multi-task fine-tuning.}
\centering
\resizebox{\textwidth}{!}{%
\begin{tabular}{|l|l|l|}
\hline
\textbf{Dataset Name} & \textbf{Domain} & \textbf{Number of Training Samples} \\ \hline
MetaMathQA~\cite{yu2023metamath} & Mathematics & 395K \\ \hline
CodeFeedback~\cite{zheng2024opencodeinterpreter} & Code Generation & 105K \\ \hline
WizardLM-Evol-Instruct~\cite{xu2024wizardlm} & Conversation & 143K \\ \hline
BoolQ~\cite{clark2019boolq} & yes/no QA & 9.4K \\ \hline
PIQA~\cite{bisk2020piqa} & physical commonsense & 16.1K \\ \hline
SIQA~\cite{sap2019socialiqa} & social reasoning & 33.4K \\ \hline
HellaSwag~\cite{zellers2019hellaswag} & commonsense NLI & 39.9K \\ \hline
WinoGrande~\cite{sakaguchi2021winogrande}  & fill-in-the-blank & 40.4K \\ \hline
ARC\_Challenge~\cite{clark2018think} & multiple-choice science questions & 1.1K \\ \hline
ARC\_Easy~\cite{clark2018think} & multiple-choice science questions & 2.3K \\ \hline
OpenBookQA~\cite{mihaylov2018can} & multi-step reasoning & 5.0K \\ \hline
\end{tabular}
}
\label{tab:multi-dataset}
\end{table*}



\section{Speed Comparison}
\label{appendix:speed}

\begin{table}[ht]
\centering
\caption{Training overhead comparison between PiSSA and HD-PiSSA on LLaMA-2-7B using 8 GPUs. Both methods are run for 100 steps on the MetaMath dataset without advanced memory or acceleration optimizations.}
\label{tab:overhead_comparison}
\begin{tabular}{lcc}
\toprule
\textbf{Metric} & \textbf{PiSSA} & \textbf{HD-PiSSA} \\
\midrule
Peak Mem.  & 47,290 MiB & 49,176 MiB \\
Time (100 steps)    & 1808.31 s  & 1995.8 s \\
\bottomrule
\end{tabular}
\end{table}

To evaluate the training overhead of our proposed HD-PiSSA, we conducted a comparative analysis against PiSSA. The experiment was performed on the MetaMath dataset using a LLaMA-2-7B model, distributed across 8 GPUs with 16 ranks per GPU, and using bf16 precision without any advanced memory and acceleration optimizations (our implementation of parallelization and communication). 

As detailed in \textbf{Table~\ref{tab:overhead_comparison}}, HD-PiSSA exhibits a slight increase in both peak memory usage per GPU (49,176 MiB vs. 47,290 MiB) and total training time for 100 steps (1995.8s vs. 1808.31s). We attribute this marginal overhead to the additional communication and computation required for updating the full-parameter matrix $W$. For instance, each GPU is required to gather the previous values and gradients of matrices $A$ and $B$ from all other GPUs, followed by computing the corresponding update to $W$. Nevertheless, we consider this overhead to be acceptable given the substantial gains in performance demonstrated by HD-PiSSA.





\end{document}